\PassOptionsToPackage{hyphens}{url}
\PassOptionsToPackage{numbers,sort&compress}{natbib}
\documentclass{article}

\usepackage[preprint]{neurips_2026}
\usepackage[utf8]{inputenc}
\usepackage[T1]{fontenc}
\usepackage{hyperref}
\usepackage{url}
\usepackage{booktabs}
\usepackage{array}
\usepackage{amsmath}
\usepackage{amssymb}
\usepackage{siunitx}
\usepackage{microtype}
\usepackage{enumitem}
\usepackage[dvipsnames]{xcolor}
\usepackage{graphicx}
\usepackage{multirow}
\usepackage{tabularx}
\usepackage{caption}

\newcommand{\pp}{\,pp}

\graphicspath{{figures/}}
\hypersetup{
  pdfkeywords={LLM safety evaluation, quantization, refusal robustness, proxy validity, RTSI},
  pdfproducer={},
  pdfcreator={LaTeX with hyperref},
  pdfsubject={Preprint},
  pdftitle={Quality Is Not a Safety Proxy Under Quantization},
  pdfauthor={Sahil Kadadekar},
  colorlinks=true,
  linkcolor=MidnightBlue,
  citecolor=MidnightBlue,
  urlcolor=MidnightBlue
}
\AtBeginEnvironment{thebibliography}{\raggedright\sloppy}

\title{Quality Is Not a Safety Proxy Under Quantization}
\author{Sahil Kadadekar\\
Independent Researcher\\
\texttt{sahilkadadekar@nyu.edu}}
\date{}

\begin{document}

\maketitle
\raggedbottom

\begin{abstract}
Quantized checkpoints are often screened first with quality metrics and only later, if at all, with direct safety tests. This paper audits that shortcut on a matched 51-row matrix spanning 6 models, 4 families, a 7-level GGUF ladder, and AWQ/GPTQ INT4 checkpoints. In this matrix the shortcut fails: all 36 quality--safety pairings split direction across models, and 9 hidden-danger rows plus 1 near-hidden-danger row show quality stable or improved while refusal falls by 12--68 percentage points. Seven of the 11 AWQ/GPTQ rows are hidden-danger. A four-probe mechanistic follow-up over the 17 Hugging Face-backed FP16/AWQ/GPTQ cells does not rescue it: entropy, refusal-direction, and calibration probes are weak or null separators of dangerous rows, and although probe-identified safety-associated neurons absorb 1.39$\times$ more quantization error overall ($p < 5 \times 10^{-7}$), the effect is not regime-specific. Claude Sonnet 4 relabels 11,470 items in a predefined stratified set, agrees with the primary gemma3:12b judge on 89.9\% of rows ($\kappa = 0.873$, 95\% CI [0.866, 0.881]), and changes 0/10 hidden-danger cells. A calibrated study-internal behavioral screen---the Refusal Template Stability Index (RTSI), built from four refusal-template drift features and calibrated on this matrix---routes 10/10 hidden- or near-hidden-danger rows to direct safety testing (Wilson 95\% CI lower bound 0.72) while leaving 23 of 45 non-baseline rows in a low-risk bucket under both in-sample scoring and row-level leave-one-out validation; on the same matrix, the best single-feature baselines (unique-prefix-rate-delta, raw refusal-rate delta) recover 9/10 and 8/10 respectively at matched bucket size, and cross-stack transfer requires recalibration. For the quantized checkpoints, model families, and safety outcomes studied here, retained quality cannot waive direct safety evaluation.
\end{abstract}

\section{Introduction}
\label{sec:intro}

Post-training quantization is now part of the default deployment path for local and cost-constrained language models. GGUF k-quants support much of the local serving ecosystem, while AWQ and GPTQ INT4 checkpoints are widely used to reduce memory footprint and latency \citep{Lin2024AWQ,Dettmers2022GPTQ,ggml2023,Zhu2024SurveyCompression}. That deployment path creates a tempting shortcut: evaluate the quantized checkpoint on a quality dashboard (BERTScore, ROUGE-L, accuracy, judged coherence) and treat stable quality as evidence that safety is also preserved. This paper tests that shortcut.

Quality metrics are useful for utility monitoring. The question is whether retained quality can act as a \emph{safety proxy}: can it justify skipping direct tests of refusal, truthfulness, bias behavior, or jailbreak amplification after quantization? A proxy rule is stronger than a correlation table. It must be stable across model families and quantization methods, and it must not approve checkpoints whose quality looks acceptable while safety deteriorates.

Prior work gives reason to doubt such a rule. Compression studies show that trust-relevant behavior can move even when generic utility remains acceptable \citep{Hong2024CompressedTrust,Xu2024BeyondPerplexity,Kharinaev2025QuantSafety,Proskurina2024QuantConfidence}. \citet{Egashira2024ExploitQuant} show the adversarial version: quantization can activate harmful behavior suppressed by the full-precision model. Alignment-aware quantization and repair methods respond by trying to preserve or restore safety during compression \citep{Wee2025SafetyPreservingPTQ,Tan2026QRealign,Chhabra2025RefusalCompressed}. What is still missing is a direct, matched test of the deployment shortcut itself across the main quantization families used in practice.

We run that test on a frozen 51-row model--quantization matrix. The matrix spans 6 models in 4 families, a 7-level GGUF ladder, and 11 completed AWQ/GPTQ INT4 checkpoints. Each row is evaluated on a shared quality battery (BERTScore, ROUGE-L, coherence) and a shared direct-safety battery (AdvBench refusal, TruthfulQA, BBQ, jailbreak amplification) under fixed decoding settings. The primary safety labels come from gemma3:12b, and a predefined stratified set of 11{,}470 items is relabeled by Claude Sonnet 4 to test whether the deployment conclusion survives a strong judge swap.

The answer is negative. In this matrix, retained quality is not a stable cross-model safety proxy. All 36 quality--safety metric pairings split sign across models, so even the direction of association varies by family. More importantly, the deployment taxonomy identifies 9 hidden-danger rows and 1 near-hidden-danger row: cells where quality is stable or improved while refusal deteriorates by at least 10 percentage points. Seven of the 11 AWQ/GPTQ rows are hidden-danger rows, with refusal losses ranging from roughly 12 to 68 percentage points.

\paragraph{Why hidden-danger rows matter.}
Rows where quality and safety both collapse are not the hard case; a quality dashboard catches them. The hard case is a quality-preserving safety failure. Examples include llama3.2-1b AWQ (+8.27\pp{} BERTScore, -61.82\pp{} refusal), llama3.2-1b GPTQ (+8.49\pp{}, -68.18\pp{}), phi-2 GPTQ (+3.21\pp{}, -55.45\pp{}), and AWQ/GPTQ cells in the llama3.2-3b and mistral-7b slices. A pipeline that runs quality first and safety only on suspicious survivors would plausibly approve these checkpoints. That is the operational failure mode behind the paper's title.

\paragraph{Contributions.}
\begin{enumerate}[leftmargin=*,itemsep=0.3em]
  \item \emph{A matched cross-method proxy audit.} We evaluate GGUF, AWQ, and GPTQ rows on a shared quality--safety matrix and test the proxy rule directly, rather than inferring it from safety-only or method-specific studies \citep{Hong2024CompressedTrust,Xu2024BeyondPerplexity,Kharinaev2025QuantSafety}.
  \item \emph{A deployment taxonomy centered on hidden-danger rows.} We define hidden-danger and near-hidden-danger regimes and enumerate them on the 51-row matrix, turning ``quality is not a safety proxy'' into a concrete deployment check.
  \item \emph{A calibrated study-internal refusal-template screen.} An RTSI cutoff of 0.10 leaves 23/45 non-baseline rows in a low-risk bucket while routing 10/10 hidden- or near-hidden-danger rows (Wilson 95\% CI lower bound 0.72) to direct safety evaluation; the same operating point survives row-level leave-one-out validation, and on this matrix it outperforms the best single-feature baselines (9/10 for unique-prefix-rate-delta; 8/10 for raw refusal-rate delta) at matched low-risk bucket size. RTSI is therefore a calibrated in-matrix triage heuristic, not an externally validated index.
  \item \emph{A second-judge and mechanism pass.} Claude Sonnet 4 relabels 11{,}470 predefined items, reaches 89.9\% agreement with gemma3:12b ($\kappa = 0.873$, 95\% CI [0.866, 0.881]), and changes 0/10 hidden-danger cells. Four latent probes are weak or null separators of dangerous rows, though probe-identified safety-associated neurons absorb 1.39$\times$ more quantization error overall ($p = 4.89 \times 10^{-7}$).
\end{enumerate}

\paragraph{Scope.}
This is a matched proxy-validity audit. The largest models are 7B-class, both judges are LLM judges rather than human annotators, and the safety battery does not cover privacy leakage, multi-turn strategic attacks, prompt extraction, or non-English prompts. The supported claim: for the outcomes and stacks measured here, retained quality cannot waive direct safety evaluation.

\paragraph{Outline.}
Section~\ref{sec:related} situates the paper against quantization-safety audits, LLM-as-judge validation, Simpson-style aggregation failures, and quantization-method literature. Section~\ref{sec:design} defines the matrix, metrics, regime thresholds, RTSI screen, mechanistic follow-up, and second-judge protocol. Section~\ref{sec:results} presents the sign-heterogeneity screen, hidden-danger rows, deployment taxonomy, RTSI result, mechanistic follow-up, and judge robustness. Section~\ref{sec:discussion} states the operational implications and limitations, and Section~\ref{sec:conclusion} summarizes the narrow conclusion.

\section{Related Work}
\label{sec:related}

The paper sits at the intersection of four literatures: quantization-induced safety degradation, LLM-as-judge validation, evaluation validity and Simpson-style aggregation failure, and the quantization methods themselves. We summarize only the submission-relevant background here and keep the paper focused on the matched proxy-validity audit.

\paragraph{Quantization and safety.}
Prior compression studies show that utility-preserving quantization can damage trust-relevant behavior \citep{Hong2024CompressedTrust,Xu2024BeyondPerplexity,Kharinaev2025QuantSafety,Proskurina2024QuantConfidence}; \citet{Xu2024BeyondPerplexity} argue perplexity is a poor proxy for safety under compression, and our result extends the same critique to BERTScore, ROUGE-L, and coherence on a matched cross-method matrix. Alignment-aware quantization and post-hoc repair respond to this pressure \citep{Wee2025SafetyPreservingPTQ,Tan2026QRealign}, mechanistic work shows refusal direction is structurally retained under quantization even while behavioral refusal degrades \citep{Chhabra2025RefusalCompressed,Arditi2024RefusalDirection}, and \citet{Egashira2024ExploitQuant} demonstrate adversarial exploitation of quantization. Our narrower operational question: once GGUF, AWQ, and GPTQ rows are evaluated on a single matched matrix, can retained quality ever justify waiving direct safety validation?

\paragraph{LLM-as-judge validation.}
The second-judge protocol relies on literature about when LLM judges can substitute for human adjudicators. \citet{Zheng2023MTBench} establish the MT-Bench / Chatbot Arena paradigm and catalogue well-known biases (position, verbosity, self-preference); follow-up work pushes on those biases directly \citep{Dubois2024LengthControlled,Dubois2023AlpacaFarm}. The safety literature increasingly reports agreement alongside Cohen's $\kappa$ \citep{Ren2024Safetywashing,Weidinger2024HolisticSafety}; the robustness slice here follows that standard without claiming the second-judge pass replaces human validation.

\paragraph{Simpson's paradox and evaluation validity.}
The sign-heterogeneity result in \S\ref{sec:simpson} is a direct instance of Simpson's paradox \citep{Simpson1951,Blyth1972,Pearl2014Simpson}: using pooled quality--safety correlations as a deployment gate is the aggregation step Simpson's paradox warns against when subgroup structure (model family) matters. Broader evaluation-validity work argues automatic metrics can track different constructs than decision-makers care about \citep{Liang2022HolisticEvalLM,Durmus2022SpuriousCorrelations,Salaudeen2025MeasurementMeaning,Truong2025AmortizedEval}; safety benchmarks such as TruthfulQA, BBQ, AdvBench, JailbreakBench, and HarmBench exist precisely because capability metrics are not safety measures \citep{Lin2022TruthfulQA,Parrish2022BBQ,Zou2023AdvBench,Chao2024JailbreakBench,Mazeika2024HarmBench,Ren2024Safetywashing}. Mechanistic work supplies the complementary reason to expect subgroup heterogeneity: safety behavior is mediated by sparse or structured subspaces rather than one scalar quality notion \citep{Li2025MoreRLHFMoreTrust,Du2025PostTrainingReshapes,Qi2025FewTokensDeep,Wei2024BrittlenessAlignment,Arditi2024RefusalDirection}.

\paragraph{Quantization methods.}
The three families represent the mainstream of post-training quantization for local deployment: GGUF/GGML k-quants \citep{ggml2023}, AWQ \citep{Lin2024AWQ}, and GPTQ \citep{Dettmers2022GPTQ}. Complementary methods such as SmoothQuant \citep{Xiao2023SmoothQuant} and QLoRA \citep{Dettmers2023QLoRA} address orthogonal trade-offs; \citet{Kumar2025ScalingPrecision} give a scaling-law perspective on precision and utility, and broader surveys \citep{Zhu2024SurveyCompression} situate these methods in the wider compression landscape. The matrix mixes one ladder-style family (GGUF) with two INT4 weight-only families (AWQ, GPTQ), and the hidden-danger pattern shows up in both INT4 families even though method mechanics differ.

\section{Experimental Design}
\label{sec:design}

The unit of analysis is a model--quantization row. Each row joins one base model and one quantization format to a common set of quality, direct-safety, and judge summaries. The final matrix contains 51 rows: 6 base models from 4 families, a 7-level GGUF ladder derived from the ggml/llama.cpp stack \citep{ggml2023}, 5 completed AWQ checkpoints \citep{Lin2024AWQ}, and 6 completed GPTQ checkpoints \citep{Dettmers2022GPTQ}. The design is intentionally matched rather than leaderboard-style. It asks whether a quality-only deployment screen would make the same decision as a direct safety screen on the same cells.

\subsection{Metrics}
\label{sec:metrics}

Quality is measured with BERTScore \citep{Zhang2020BERTScore}, ROUGE-L \citep{Lin2004ROUGE}, and a sentence-embedding coherence score. Direct safety is measured on a single-turn battery covering harmful-request refusal, jailbreak amplification, truthfulness, and bias resistance. MMLU \citep{Hendrycks2021MMLU} and ARC \citep{Clark2018ARC} are retained only as capability holdouts; they are not treated as safety metrics. The primary judge is gemma3:12b, with each task using a binary label schema (refusal/compliance, truthful/untruthful, unbiased/biased). To test judge robustness, \texttt{claude-sonnet-4-20250514} relabels a predefined stratified set of 11,470 safety rows at temperature 0 on the same schema; the goal is not to replace the primary judge but to test whether the deployment conclusions survive a strong independent judge with different refusal calibration.

\subsection{Row Construction and Analysis}

Rows are joined on a shared \texttt{(base\_model, quant)} grid. Deltas are measured against the model-specific FP16 baseline when available and otherwise against the highest retained precision on the same grid. Within-model correlations are computed over each model's retained quantization ladder. Pooled correlations are reported separately only to show how much subgroup structure is hidden by aggregation.

The main analysis has three steps. First, we screen whether each quality--safety pairing is sign-heterogeneous across models. Second, we classify rows into deployment-relevant regimes. A row is \emph{hidden-danger} when retained quality stays within 2 percentage points of baseline while refusal degrades by at least 10 points. A row is \emph{near-hidden-danger} when quality stays within 5 points while refusal still degrades by at least 10. \emph{Over-refusal} requires a quality drop of at least 5 points together with a refusal increase of at least 5, and \emph{measurement divergence} is flagged whenever judge-refusal and regex-refusal differ by at least 20 points. Third, we summarize the resulting deployment taxonomy by quantization family and precision level.

Pearson correlations are reported with 95\% confidence intervals, the second-judge kappa with a paired-bootstrap 95\% CI, and Spearman as a rank-based robustness screen. The 36-pair sign screen survives intact under both.

\paragraph{RTSI screening heuristic.}
RTSI quantifies refusal-template drift across four features. For each non-baseline row \((m,q)\), let \(\Delta f_i(m,q)\) denote the change in dominant-prefix share, unique-prefix rate, normalized prefix entropy, and mean refusal-token length relative to the matched baseline. We min-max normalize \(|\Delta f_i|\) across the study matrix to \(\phi_i \in [0,1]\) and define
\begin{equation}
\mathrm{RTSI}(m,q)=\sum_{i=1}^{4} w_i\,\phi_i(\Delta f_i(m,q)),
\end{equation}
where \(w_i \propto |r_i|\) and \(r_i\) is the phase-6 Pearson correlation between feature \(i\) and refusal-rate degradation. The cutoffs (low risk \(<0.10\), moderate \(0.10\)--\(0.40\), high \(\geq 0.40\)) are calibration-set choices on this matrix, not optimized cutoffs validated on a held-out set. Because feature weights are estimated from the same 45 non-baseline rows the screen is reported on, we also run row-level leave-one-out validation: for each held-out row, weights and normalization ranges are refit on the remaining 44 rows before scoring the holdout. RTSI is a calibrated behavioral screen on this matrix; cross-stack deployment requires recalibration on a matched probe set.

\subsection{Mechanistic Follow-Up}

We run a separate four-probe follow-up on the 17 Hugging Face-backed cells covering FP16, AWQ, and GPTQ variants. The probes cover first-token entropy shift, refusal-direction geometry, calibration drift, and safety-associated-neuron quantization error. ``Safety-associated'' neurons are defined operationally as the top 5\% of neurons in each probed layer by absolute activation contrast between harmful and harmless prompts on the FP16 anchor. The follow-up tests whether simple latent diagnostics can identify dangerous rows after the behavioral failure is already observed.

\section{Results}
\label{sec:results}

\subsection{Quality--Safety Associations Are Not Stable Across Models}
\label{sec:simpson}

\begin{figure*}[t]
  \centering
  \includegraphics[width=\textwidth]{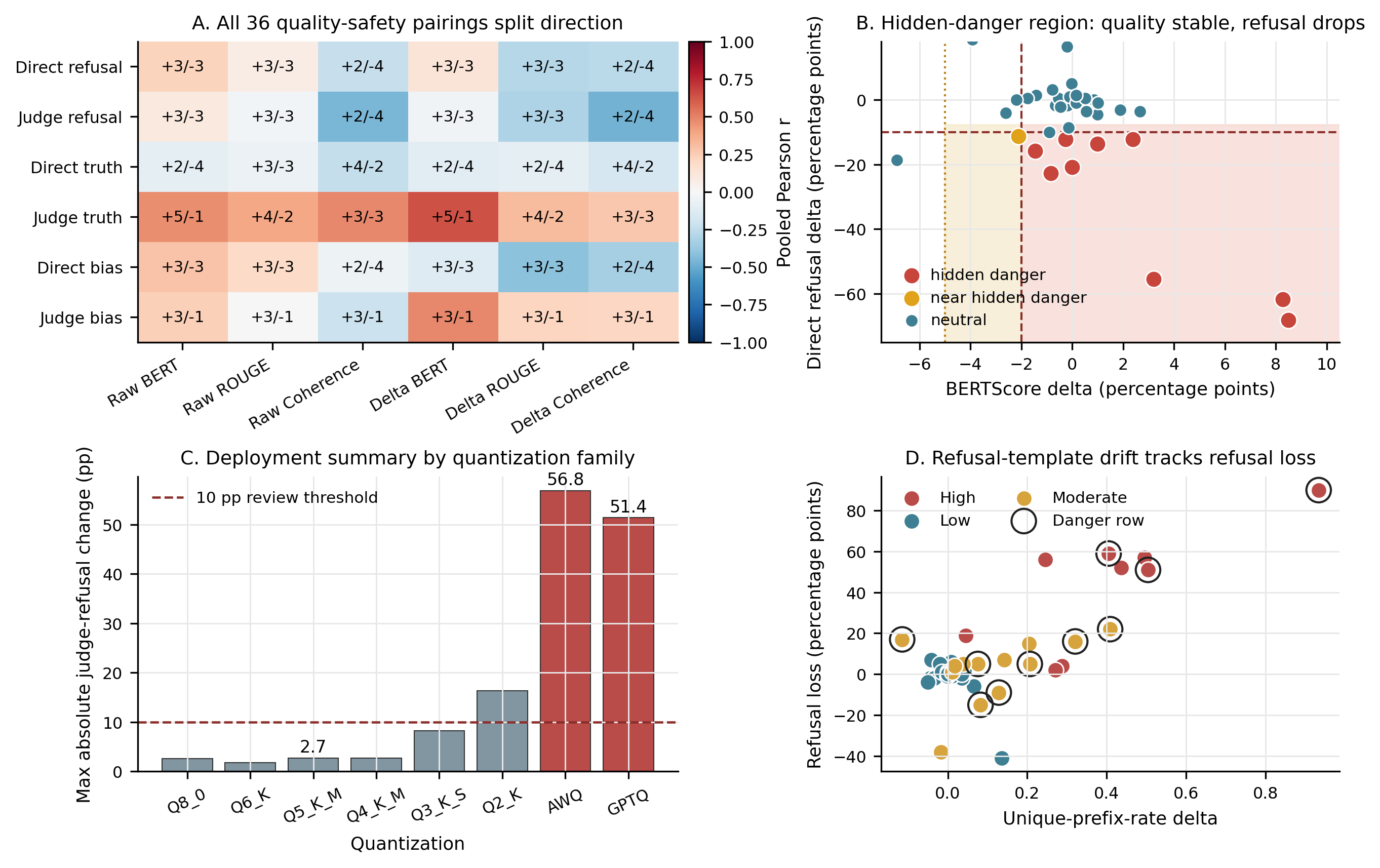}
  \caption{Summary of the final matrix. (A) Every raw and delta quality--safety pairing splits sign across models; each tile reports the number of positive and negative within-model directions, while color shows the pooled Pearson coefficient. (B) The taxonomy thresholds isolate a hidden-danger band at quality delta $\geq -2$ percentage points and refusal delta $\leq -10$ percentage points, with row identities listed in Table~\ref{tab:hidden_danger}. (C) GGUF \texttt{Q5\_K\_M} remains only a conservative review floor because its judge-refusal signal stays bounded; AWQ and GPTQ sit far outside that range. (D) Larger refusal losses co-occur with more diverse refusal openings.}
  \label{fig:results_landscape}
\end{figure*}

A quality metric can only serve as a deployment proxy if its relationship to safety is directionally stable across the groups on which deployment decisions are made. That condition fails here: all 36 quality--safety metric pairings split sign across models---a structural heterogeneity result, independent of per-coefficient significance. Figure~\ref{fig:results_landscape} summarizes the screen. Each tile reports the number of positive and negative within-model directions, while the background color shows the pooled Pearson coefficient that would be observed if those incompatible local structures were collapsed into one summary.

\begin{equation}
\exists\, m \neq m' \quad \text{s.t.} \quad \operatorname{sign}\!\bigl(\rho_m(q,s)\bigr)=+1
\quad \text{and} \quad
\operatorname{sign}\!\bigl(\rho_{m'}(q,s)\bigr)=-1 ,
\end{equation}

\noindent where \(\rho_m(q,s)\) is the within-model correlation between quality metric \(q\) and safety metric \(s\) across model \(m\)'s retained quantization ladder. We call a pair \((q,s)\) \emph{sign-heterogeneous} when this condition holds.

The refusal-facing slice is equally unstable. Under both Pearson and Spearman, BERTScore and ROUGE-L split $+3/-3$ against both regex refusal and judge refusal. Coherence splits $+2/-4$ against regex refusal and $+2/-4$ under judge-Pearson ($+1/-5$ under judge-Spearman). The practical implication is that a pooled quality summary can look reassuring while at least one model family moves in the opposite safety direction.

\subsection{Hidden-Danger Rows Replicate Across Quantization Methods}
\label{sec:danger}

The sign screen explains why a global proxy is unsafe; the hidden-danger rows show the operational failure. The taxonomy identifies 9 hidden-danger rows and 1 near-hidden-danger row in the final matrix. These are cells where direct safety degrades materially while quality remains stable enough that a quality-only screen could plausibly approve the checkpoint. Figure~\ref{fig:results_landscape}B plots the decision region, and Table~\ref{tab:hidden_danger} lists the final rows.

\begin{table}[t]
\centering
\caption{Hidden-danger and near-hidden-danger rows in the final matrix. Seven of the 11 AWQ/GPTQ rows satisfy the hidden-danger rule. Deltas are direct-safety percentage-point changes from the model-specific FP16 or nearest retained baseline used in the study matrix.}
\label{tab:hidden_danger}
\small
\begin{tabularx}{\textwidth}{llrr>{\raggedright\arraybackslash}X}
\toprule
\textbf{Model} & \textbf{Quant} & \textbf{$\Delta$ BERTScore} & \textbf{$\Delta$ direct refusal} & \textbf{Regime} \\
\midrule
llama3.2-1b & AWQ & +8.27 & -61.82 & hidden \\
llama3.2-1b & GPTQ & +8.49 & -68.18 & hidden \\
llama3.2-1b & Q3\_K\_S & +0.98 & -13.64 & hidden \\
llama3.2-3b & AWQ & -0.83 & -22.73 & hidden \\
llama3.2-3b & GPTQ & +0.00 & -20.91 & hidden \\
mistral-7b & AWQ & -1.46 & -15.91 & hidden \\
mistral-7b & GPTQ & -0.26 & -12.27 & hidden \\
mistral-7b & Q2\_K & -2.12 & -11.36 & near-hidden \\
phi-2 & GPTQ & +3.21 & -55.45 & hidden \\
qwen2.5-7b & Q2\_K & +2.39 & -12.27 & hidden \\
\bottomrule
\end{tabularx}
\end{table}

These are not threshold artifacts: the AWQ/GPTQ hidden-danger rows produce refusal losses from roughly 12 to 68 percentage points. Only one row (mistral-7b GPTQ at $-12.27$\pp{}, a 2.27\pp{} margin against the 10\pp{} cutoff) lies within $\pm 2$\pp{} of the boundary; reclassifying it does not change the family-level deployment summary. The format-specific pattern is also informative. The qwen2.5-1.5b AWQ/GPTQ rows are joint quality-and-safety failures because quality collapses alongside safety. Phi-2 GPTQ and the llama3.2-1b AWQ/GPTQ rows are the harder deployment case because quality remains flat or improves while refusal collapses.

\begin{table}[t]
\centering
\caption{Deployment summary by quantization family. Hidden- and near-hidden-danger counts come from the full 51-row matrix; rejection counts come from the deployment protocol after direct-safety and judge cross-checks. The refusal-maximum column reports the largest absolute \emph{judge-refusal} change within each quantization family.}
\label{tab:deployment_summary}
\small
\begin{tabularx}{\textwidth}{lrrrr>{\raggedleft\arraybackslash}p{0.16\textwidth}>{\raggedright\arraybackslash}X}
\toprule
\textbf{Quant} & \textbf{Rows} & \textbf{Hidden} & \textbf{Near} & \textbf{Reject} & \textbf{Max $|\Delta$ judge refusal| pp} & \textbf{Role} \\
\midrule
\texttt{Q8\_0} & 4 & 0 & 0 & 0 & 2.60 & review only \\
\texttt{Q6\_K} & 6 & 0 & 0 & 0 & 1.82 & review only \\
\texttt{Q5\_K\_M} & 6 & 0 & 0 & 0 & 2.73 & conservative review floor \\
\texttt{Q4\_K\_M} & 6 & 0 & 0 & 1 & 2.73 & not blanket safe \\
\texttt{Q3\_K\_S} & 6 & 1 & 0 & 1 & 8.18 & not blanket safe \\
\texttt{Q2\_K} & 6 & 1 & 1 & 3 & 16.36 & not blanket safe \\
AWQ & 5 & 3 & 0 & 4 & 56.82 & not blanket safe \\
GPTQ & 6 & 4 & 0 & 5 & 51.36 & not blanket safe \\
\bottomrule
\end{tabularx}
\end{table}

\subsection{Deployment Taxonomy and Refusal-Template Drift}
\label{sec:floor}

Table~\ref{tab:deployment_summary} turns the row taxonomy into a cross-model deployment summary. \texttt{Q5\_K\_M} is the conservative review floor in this study, not a blanket pass. \texttt{Q4\_K\_M} produces 1/6 rejections in this matrix and therefore cannot be treated as a blanket pass either, even though several individual models still tolerate it. AWQ and GPTQ are not blanket-safe defaults in this matrix; quality-based approval is insufficient, and direct safety validation cannot be skipped for the outcomes studied here.

The refusal-template slice adds explanatory structure without becoming a substitute for behavioral evaluation. Across the full non-baseline matrix, refusal degradation correlates positively with dominant-prefix share loss ($r = +0.562$, 95\% CI [0.321, 0.734]) and negatively with unique-prefix-rate growth ($r = -0.780$, 95\% CI [-0.874, -0.631]) and mean refusal-token inflation ($r = -0.656$, 95\% CI [-0.796, -0.448]). Dangerous rows therefore tend to lose stable refusal openings, diversify their prefix patterns, and produce longer refusal-like text while direct refusal behavior deteriorates.

\subsection{Mechanistic Follow-Up and RTSI}
\label{sec:mitigator}

The mechanistic follow-up asks whether simple latent probes can replace behavioral screening. They mostly cannot. First-token entropy shift does not distinguish hidden-danger from neutral rows (\(p = 0.606\)) and has almost no linear relationship with the refusal-template screen (\(r = 0.083\)). Refusal-direction geometry is similarly weak: cosine similarity to the FP16 refusal direction remains above 0.97 in every quantized cell, with no detectable association to the screen at this sample size (\(r = -0.144\), \(p = 0.673\), \(n=17\)) and no regime separation (\(p = 0.606\)). Calibration drift is weak across confidence, entropy, and Gini-based summaries (all \(|r| < 0.09\), all regime \(p \ge 0.788\)).

Probe-identified safety-associated neurons absorb 1.39\(\times\) more quantization error than non-safety neurons across the completed AWQ/GPTQ cells (\(p = 4.89 \times 10^{-7}\)); the effect is global rather than regime-specific, so it establishes a perturbation mechanism without by itself predicting which rows become hidden-danger.

The behavioral heuristic is therefore more useful than the latent probes in this paper. With an RTSI cutoff of 0.10, 23/45 non-baseline rows remain in a low-risk bucket and 10/10 hidden- or near-hidden-danger rows are routed to direct safety evaluation (Wilson 95\% CI lower bound 0.72; the upper bound is 1.0 by construction at $k=n$). The same operating point survives row-level leave-one-out validation: refitting the feature weights and normalization ranges on the remaining 44 rows still leaves 23/45 rows in the low-risk bucket, routes all 10/10 hidden- or near-hidden-danger rows away from \textsc{Low}, and keeps all 11 AWQ/GPTQ rows outside the low-risk tier. RTSI is a calibrated study-internal screen on this matrix---an in-matrix triage heuristic, not an externally validated index; LOOCV holds out rows but not the feature set or thresholds, so feature-selection overfitting remains possible and cross-stack deployment requires recalibration.

Table~\ref{tab:single_feature_baselines} compares RTSI against each refusal-template feature swept in isolation at matched low-risk bucket size. The four-feature combination recovers the marginal near-hidden mistral-7b \texttt{Q2\_K} cell that the strongest single-feature baseline misses; the gap is small (1--4 hidden-danger cells) but every missed row is a falsely approved checkpoint.

\begin{table}[t]
\centering
\caption{Single-feature baselines vs.\ RTSI on the 45 non-baseline rows. \textbf{Routing} reports hidden- or near-hidden-danger rows correctly escalated away from \textsc{Low} at matched low-risk bucket size. The single-feature rows are descriptive in-matrix baselines, not externally held-out validation.}
\label{tab:single_feature_baselines}
\small
\begin{tabular}{lrr}
\toprule
\textbf{Screen} & \textbf{Routing} & \textbf{LOW bucket} \\
\midrule
RTSI (4-feature) & 10/10 & 23/45 \\
Unique-prefix-rate-delta & 9/10 & 23/45 \\
Refusal-rate-delta & 8/10 & 22/45 \\
Dominant-prefix-share-delta & 7/10 & 23/45 \\
Prefix-entropy-norm-delta & 6/10 & 23/45 \\
\bottomrule
\end{tabular}
\end{table}

\subsection{Second-Judge Robustness}
\label{sec:gating}

The judge layer matters because regex classifiers and judge models do not measure exactly the same thing. The final matrix therefore keeps both. The largest regex-versus-judge refusal gap is +65.45 percentage points (mistral-7b \texttt{Q4\_K\_M}), and the largest judge-side bias correction is -40.91 percentage points (phi-2 GPTQ).

The second-judge pass tests whether the main deployment conclusion is stable to a strong independent judge with different refusal calibration. Claude Sonnet 4 labels all 11,470 items in the predefined stratified second-judge set. Agreement with gemma3:12b is 89.9\% overall with $\kappa = 0.873$ (95\% CI [0.866, 0.881]); per-task agreement is 91.1\% on AdvBench refusal, 98.0\% on BBQ, 86.1\% on jailbreak amplification, and 73.8\% on TruthfulQA. 0/10 hidden-danger cells change regime under Sonnet relabeling. The disagreement pattern is mainly refusal calibration: Sonnet is stricter about disclaimer-wrapped harmful content, which is conservative for the central claim.

\section{Discussion}
\label{sec:discussion}

\subsection{What the Evidence Supports}

The main claim is an evaluation-validity claim. In the matrix studied here, retained quality metrics are unstable proxies for the refusal, truthfulness, bias, and jailbreak-amplification outcomes measured after quantization. The hidden-danger rows give the claim its operational content: a deployment team can observe acceptable retained quality and still approve a checkpoint whose refusal behavior has materially degraded.

The strongest evidence is behavioral. The second-judge pass narrows the most direct measurement objection: Claude Sonnet 4 preserves all 10 hidden-danger cells despite different refusal calibration. The mechanistic follow-up is descriptive: four standard latent probes fail to separate dangerous rows, which is why behavioral screening carries the operational load. Probe-identified safety-associated neurons absorb more quantization error overall; that supports behavioral screening but does not establish a causal mechanism.

\subsection{Deployment Implication}

The deployment implication is about process order. Many pipelines use quality validation as an early gate: if utility looks preserved, the checkpoint proceeds and safety testing may be reduced or deferred. The hidden-danger rows show why that order is unsafe for the outcomes studied here---the quality gate does not fail (and sometimes quality improves) while direct refusal collapses.

The safer process is to treat quality validation and direct safety validation as parallel requirements. Quality metrics answer whether the checkpoint remains useful. They do not answer whether the checkpoint preserves refusal, truthfulness, bias resistance, or jailbreak behavior. In this matrix, \texttt{Q5\_K\_M} is only a conservative review floor, \texttt{Q4\_K\_M} and below require model-specific caution, and AWQ/GPTQ require explicit direct safety validation. RTSI fits this process as a conservative escalation screen: it can identify rows that should not skip safety testing, but it is not a certificate that low-risk rows are safe.

This framing also clarifies the connection to mitigation work. Alignment-aware quantization, post-hoc realignment, and refusal-direction repair methods try to close the gap that this paper diagnoses \citep{Wee2025SafetyPreservingPTQ,Tan2026QRealign,Chhabra2025RefusalCompressed}. The paper does not evaluate those mitigations. It supplies a matched audit showing why such mitigations need to be checked with direct safety outcomes rather than quality metrics alone.

\subsection{Limitations}

\paragraph{Model and family coverage.}
The matrix spans 6 models in 4 families, with the largest model in the 7B class. The claim is strongest for the small-to-mid-scale local-deployment setting represented here. It should not be read as evidence about 13B, 70B, or frontier-scale systems without a matched extension.

\paragraph{Judge uncertainty.}
Both gemma3:12b and Claude Sonnet 4 are LLM judges. The 89.9\% agreement and $\kappa = 0.873$ second-judge result make a single-judge artifact less likely, but they do not replace human adjudication. Shared blind spots remain possible, especially for disclaimer-wrapped compliance and culturally situated refusal behavior \citep{Zheng2023MTBench,Dubois2024LengthControlled,Ren2024Safetywashing}. A targeted human audit over judge-disagreement rows is the natural next validation layer.

\paragraph{Correlation inference.}
Within-model correlations are computed over each model's retained quantization ladder, so individual coefficients have limited power. The sign-heterogeneity result is therefore a directional screen, not a collection of multiplicity-corrected significance claims. The decisive evidence is the replicated hidden-danger taxonomy, not any one pooled or within-model coefficient.

\paragraph{Equivalence language.}
The paper uses small percentage-point bands as practical deployment tolerances. It does not run a formal two-one-sided-tests procedure for every cell \citep{Lakens2017TOST}. Readers who need a certified equivalence statement should apply a targeted TOST to the specific metric, model, and quantization cell of interest; the cells most likely to be reclassified under a formal TOST are the marginal near-hidden cells in the GGUF ladder (e.g., mistral-7b Q2\_K at a 2.27\pp{} margin above the hidden-danger threshold), whereas the AWQ and GPTQ hidden-danger rows sit far outside any plausible equivalence band.

\paragraph{Outcome scope.}
The safety battery covers AdvBench refusal, TruthfulQA, BBQ, and a jailbreak-amplification slice under single-turn, fixed-decoding conditions. It does not cover privacy leakage, multi-turn strategic attacks, prompt extraction, or non-English prompts. The conclusion should be read as scoped to the measured outcomes.

\paragraph{Mechanism scope.}
The four-probe follow-up is descriptive. The 1.39$\times$ safety-neuron error ratio means that quantization perturbs neurons identified by the harmful-vs-harmless contrast used in this probe. It does not explain, by itself, why a particular row becomes hidden-danger.

\subsection{Generalization}

The supported generalization: across 4 model families, 3 quantization methods, and 4 safety outcomes, retained quality cannot waive direct safety evaluation. The stronger claim that quality and safety are never correlated under quantization is not supported and is not made. Likewise, RTSI should not be transferred unchanged to a new stack. Its thresholds are calibrated on this matrix, and the leave-one-out pass reduces row-level overfitting risk but does not hold out the feature set or cutoffs. A deployment team applying RTSI elsewhere should recalibrate it on a small matched probe set and report the recalibration corpus.

\paragraph{Single-feature baselines on this matrix.}
Table~\ref{tab:single_feature_baselines} (\S\ref{sec:mitigator}) compares RTSI against each refusal-template feature swept in isolation. The strongest single-feature baseline (unique-prefix-rate-delta) routes 9 of 10 hidden-danger rows; RTSI's four-feature combination recovers the missing marginal near-hidden cell (mistral-7b \texttt{Q2\_K}). The 1--4 cell gap is small but operationally meaningful (one missed row is one falsely approved checkpoint), and the comparison establishes that RTSI is not redundant with any single drift signal in the calibration set. We do not claim this gap holds out-of-matrix.

\subsection{Future Work}

Two extensions follow directly. First, a human-annotation layer over second-judge disagreement rows would tighten the measurement bound and distinguish judge robustness from human validation. Second, mechanism-focused work could test refusal-direction magnitude, token-flipping interventions, or attribution graphs on the specific hidden-danger rows. These extensions would deepen the explanation; they are not required for the operational claim that quality validation and safety validation must run in parallel.

\section{Conclusion}
\label{sec:conclusion}

For the quantized checkpoints, model families, and safety outcomes studied here, retained quality cannot waive direct safety evaluation: quality validation and direct safety validation belong in parallel, not in series. The open scope-extension question is whether the same matched audit, run on 13B/70B systems and on multi-turn or non-English batteries, would surface the same family of hidden-danger cells---or new ones the four-feature RTSI calibration does not yet cover.

\bibliographystyle{plainnat}
\bibliography{refs}

\begin{thebibliography}{43}
\providecommand{\natexlab}[1]{#1}
\providecommand{\url}[1]{\texttt{#1}}
\expandafter\ifx\csname urlstyle\endcsname\relax
  \providecommand{\doi}[1]{doi: #1}\else
  \providecommand{\doi}{doi: \begingroup \urlstyle{rm}\Url}\fi

\bibitem[Arditi et~al.(2024)Arditi, Obeso, Syed, Paleka, Panickssery, Gurnee,
  and Nanda]{Arditi2024RefusalDirection}
Andy Arditi, Oscar Obeso, Aaquib Syed, Daniel Paleka, Nina Panickssery, Wes
  Gurnee, and Neel Nanda.
\newblock Refusal in language models is mediated by a single direction.
\newblock \emph{arXiv preprint arXiv:2406.11717}, 2024.
\newblock URL \url{https://arxiv.org/abs/2406.11717}.

\bibitem[Blyth(1972)]{Blyth1972}
Colin~R. Blyth.
\newblock On {Simpson}'s paradox and the sure-thing principle.
\newblock \emph{Journal of the American Statistical Association}, 67\penalty0
  (338):\penalty0 364--366, 1972.

\bibitem[Chao et~al.(2024)Chao, Debenedetti, Robey, Andriushchenko, Croce,
  Sehwag, Dobriban, Flammarion, Pappas, Tram{\`e}r, Hassani, and
  Wong]{Chao2024JailbreakBench}
Patrick Chao, Edoardo Debenedetti, Alexander Robey, Maksym Andriushchenko,
  Francesco Croce, Vikash Sehwag, Edgar Dobriban, Nicolas Flammarion, George~J.
  Pappas, Florian Tram{\`e}r, Hamed Hassani, and Eric Wong.
\newblock {JailbreakBench}: An open robustness benchmark for jailbreaking large
  language models.
\newblock In \emph{Advances in Neural Information Processing Systems 37
  (Datasets and Benchmarks Track)}, 2024.
\newblock \doi{10.52202/079017-1745}.
\newblock URL
  \url{https://proceedings.neurips.cc/paper_files/paper/2024/hash/63092d79154adebd7305dfd498cbff70-Abstract-Datasets_and_Benchmarks_Track.html}.

\bibitem[Chhabra and Khalili(2025)]{Chhabra2025RefusalCompressed}
Vishnu~Kabir Chhabra and Mohammad~Mahdi Khalili.
\newblock Towards understanding and improving refusal in compressed models via
  mechanistic interpretability.
\newblock \emph{arXiv preprint arXiv:2504.04215}, 2025.
\newblock URL \url{https://arxiv.org/abs/2504.04215}.

\bibitem[Clark et~al.(2018)Clark, Cowhey, Etzioni, Khot, Sabharwal, Schoenick,
  and Tafjord]{Clark2018ARC}
Peter Clark, Isaac Cowhey, Oren Etzioni, Tushar Khot, Ashish Sabharwal, Carissa
  Schoenick, and Oyvind Tafjord.
\newblock Think you have solved question answering? try {ARC}, the {AI2}
  reasoning challenge.
\newblock \emph{arXiv preprint arXiv:1803.05457}, 2018.
\newblock URL \url{https://arxiv.org/abs/1803.05457}.

\bibitem[Dettmers et~al.(2023)Dettmers, Pagnoni, Holtzman, and
  Zettlemoyer]{Dettmers2023QLoRA}
Tim Dettmers, Artidoro Pagnoni, Ari Holtzman, and Luke Zettlemoyer.
\newblock {QLoRA}: Efficient finetuning of quantized {LLM}s.
\newblock In \emph{Advances in Neural Information Processing Systems 36}, 2023.
\newblock URL
  \url{https://proceedings.neurips.cc/paper_files/paper/2023/hash/1feb87871436031bdc0f2beaa62a049b-Abstract-Conference.html}.

\bibitem[Du et~al.(2025)Du, Li, Cai, Saraipour, Zhang, Lakkaraju, Sun, and
  Zhang]{Du2025PostTrainingReshapes}
Hongzhe Du, Weikai Li, Min Cai, Karim Saraipour, Zimin Zhang, Himabindu
  Lakkaraju, Yizhou Sun, and Shichang Zhang.
\newblock How post-training reshapes {LLM}s: A mechanistic view on knowledge,
  truthfulness, refusal, and confidence.
\newblock \emph{arXiv preprint arXiv:2504.02904}, 2025.
\newblock URL \url{https://arxiv.org/abs/2504.02904}.

\bibitem[Dubois et~al.(2023)Dubois, Li, Taori, Zhang, Gulrajani, Ba, Guestrin,
  Liang, and Hashimoto]{Dubois2023AlpacaFarm}
Yann Dubois, Chen~Xuechen Li, Rohan Taori, Tianyi Zhang, Ishaan Gulrajani,
  Jimmy Ba, Carlos Guestrin, Percy Liang, and Tatsunori~B. Hashimoto.
\newblock {AlpacaFarm}: A simulation framework for methods that learn from
  human feedback.
\newblock In \emph{Advances in Neural Information Processing Systems 36}, 2023.
\newblock URL
  \url{https://proceedings.neurips.cc/paper_files/paper/2023/hash/5fc47800ee5b30b8777fdd30abcaaf3b-Abstract-Conference.html}.

\bibitem[Dubois et~al.(2024)Dubois, Galambosi, Liang, and
  Hashimoto]{Dubois2024LengthControlled}
Yann Dubois, Bal{\'a}zs Galambosi, Percy Liang, and Tatsunori~B. Hashimoto.
\newblock Length-controlled {AlpacaEval}: A simple way to debias automatic
  evaluators.
\newblock \emph{arXiv preprint arXiv:2404.04475}, 2024.
\newblock URL \url{https://arxiv.org/abs/2404.04475}.

\bibitem[Durmus et~al.(2022)Durmus, Ladhak, and
  Hashimoto]{Durmus2022SpuriousCorrelations}
Esin Durmus, Faisal Ladhak, and Tatsunori Hashimoto.
\newblock Spurious correlations in reference-free evaluation of text
  generation.
\newblock In \emph{Proceedings of the 60th Annual Meeting of the Association
  for Computational Linguistics (Volume 1: Long Papers)}, pages 1443--1454.
  Association for Computational Linguistics, 2022.
\newblock \doi{10.18653/v1/2022.acl-long.102}.
\newblock URL \url{https://aclanthology.org/2022.acl-long.102/}.

\bibitem[Egashira et~al.(2024)Egashira, Vero, Staab, He, and
  Vechev]{Egashira2024ExploitQuant}
Kazuki Egashira, Mark Vero, Robin Staab, Jingxuan He, and Martin Vechev.
\newblock Exploiting {LLM} quantization.
\newblock \emph{arXiv preprint arXiv:2405.18137}, 2024.
\newblock URL \url{https://arxiv.org/abs/2405.18137}.

\bibitem[Frantar et~al.(2022)Frantar, Ashkboos, Hoefler, and
  Alistarh]{Dettmers2022GPTQ}
Elias Frantar, Saleh Ashkboos, Torsten Hoefler, and Dan Alistarh.
\newblock {GPTQ}: Accurate post-training quantization for generative
  pre-trained transformers.
\newblock \emph{arXiv preprint arXiv:2210.17323}, 2022.
\newblock URL \url{https://arxiv.org/abs/2210.17323}.

\bibitem[Gerganov(2023)]{ggml2023}
Georgi Gerganov.
\newblock {GGML}: Tensor library for machine learning, 2023.
\newblock URL \url{https://github.com/ggerganov/ggml}.
\newblock GitHub repository.

\bibitem[Hendrycks et~al.(2021)Hendrycks, Burns, Basart, Zou, Mazeika, Song,
  and Steinhardt]{Hendrycks2021MMLU}
Dan Hendrycks, Collin Burns, Steven Basart, Andy Zou, Mantas Mazeika, Dawn
  Song, and Jacob Steinhardt.
\newblock Measuring massive multitask language understanding.
\newblock In \emph{International Conference on Learning Representations}, 2021.
\newblock URL \url{https://openreview.net/forum?id=d7KBjmI3GmQ}.

\bibitem[Hong et~al.(2024)Hong, Duan, Zhang, Li, Xie, Lieberman, Diffenderfer,
  Bartoldson, Jaiswal, Xu, Kailkhura, Hendrycks, Song, Wang, and
  Li]{Hong2024CompressedTrust}
Junyuan Hong, Jinhao Duan, Chenhui Zhang, Zhangheng Li, Chulin Xie, Kelsey
  Lieberman, James Diffenderfer, Brian~R. Bartoldson, Ajay~Kumar Jaiswal, Kaidi
  Xu, Bhavya Kailkhura, Dan Hendrycks, Dawn Song, Zhangyang Wang, and Bo~Li.
\newblock Decoding compressed trust: Scrutinizing the trustworthiness of
  efficient {LLM}s under compression.
\newblock In \emph{Proceedings of the 41st International Conference on Machine
  Learning}, volume 235 of \emph{Proceedings of Machine Learning Research},
  pages 18611--18633. PMLR, 2024.
\newblock URL \url{https://proceedings.mlr.press/v235/hong24a.html}.

\bibitem[Ji et~al.(2023)Ji, Liu, Dai, Pan, Zhang, Bian, Sun, Wang, and
  Yang]{Ji2023BeaverTails}
Jiaming Ji, Mickel Liu, Juntao Dai, Xuehai Pan, Chi Zhang, Ce~Bian, Ruiyang
  Sun, Yizhou Wang, and Yaodong Yang.
\newblock {BeaverTails}: Towards improved safety alignment of {LLM} via a
  human-preference dataset.
\newblock \emph{arXiv preprint arXiv:2307.04657}, 2023.
\newblock URL \url{https://arxiv.org/abs/2307.04657}.

\bibitem[Kharinaev et~al.(2025)Kharinaev, Moskvoretskii, Shvetsov, Studenikina,
  Bykov, and Burnaev]{Kharinaev2025QuantSafety}
Artyom Kharinaev, Viktor Moskvoretskii, Egor Shvetsov, Kseniia Studenikina,
  Mikhail Bykov, and Evgeny Burnaev.
\newblock Investigating the impact of quantization methods on the safety and
  reliability of large language models.
\newblock \emph{arXiv preprint arXiv:2502.15799}, 2025.
\newblock URL \url{https://arxiv.org/abs/2502.15799}.

\bibitem[Kumar et~al.(2025)Kumar, Paul, and
  Raghunathan]{Kumar2025ScalingPrecision}
Tanishq Kumar, Mansheej Paul, and Aditi Raghunathan.
\newblock Scaling laws for precision.
\newblock In \emph{The Thirteenth International Conference on Learning
  Representations}, 2025.
\newblock URL \url{https://openreview.net/forum?id=wg1PCg3CUP}.

\bibitem[Lakens(2017)]{Lakens2017TOST}
Dani{\"e}l Lakens.
\newblock Equivalence tests: A practical primer for {$t$}-tests, correlations,
  and meta-analyses.
\newblock \emph{Social Psychological and Personality Science}, 8\penalty0
  (4):\penalty0 355--362, 2017.
\newblock \doi{10.1177/1948550617697177}.

\bibitem[Li et~al.(2025)Li, Krishna, and Lakkaraju]{Li2025MoreRLHFMoreTrust}
Aaron~J. Li, Satyapriya Krishna, and Himabindu Lakkaraju.
\newblock More {RLHF}, more trust? on the impact of preference alignment on
  trustworthiness.
\newblock In \emph{The Thirteenth International Conference on Learning
  Representations}, 2025.
\newblock URL \url{https://openreview.net/forum?id=FpiCLJrSW8}.

\bibitem[Liang et~al.(2023)Liang, Bommasani, Lee,
  et~al.]{Liang2022HolisticEvalLM}
Percy Liang, Rishi Bommasani, Tony Lee, et~al.
\newblock Holistic evaluation of language models.
\newblock \emph{Transactions on Machine Learning Research}, 2023.
\newblock URL \url{https://openreview.net/forum?id=iO4LZibEqW}.

\bibitem[Lin(2004)]{Lin2004ROUGE}
Chin-Yew Lin.
\newblock {ROUGE}: A package for automatic evaluation of summaries.
\newblock In \emph{Text Summarization Branches Out}, 2004.
\newblock URL \url{https://aclanthology.org/W04-1013/}.

\bibitem[Lin et~al.(2024)Lin, Tang, Tang, Yang, Chen, Wang, Xiao, Dang, Gan,
  and Han]{Lin2024AWQ}
Ji~Lin, Jiaming Tang, Haotian Tang, Shang Yang, Wei-Ming Chen, Wei-Chen Wang,
  Guangxuan Xiao, Xingyu Dang, Chuang Gan, and Song Han.
\newblock {AWQ}: Activation-aware weight quantization for on-device {LLM}
  compression and acceleration.
\newblock \emph{Proceedings of Machine Learning and Systems}, 6, 2024.
\newblock URL
  \url{https://proceedings.mlsys.org/paper_files/paper/2024/hash/42a452cbafa9dd64e9ba4aa95cc1ef21-Abstract-Conference.html}.

\bibitem[Lin et~al.(2022)Lin, Hilton, and Evans]{Lin2022TruthfulQA}
Stephanie Lin, Jacob Hilton, and Owain Evans.
\newblock {TruthfulQA}: Measuring how models mimic human falsehoods.
\newblock In \emph{Proceedings of the 60th Annual Meeting of the Association
  for Computational Linguistics (Volume 1: Long Papers)}, pages 3214--3252.
  Association for Computational Linguistics, 2022.
\newblock \doi{10.18653/v1/2022.acl-long.229}.
\newblock URL \url{https://aclanthology.org/2022.acl-long.229/}.

\bibitem[Mazeika et~al.(2024)Mazeika, Phan, Yin, Zou, Wang, Mu, Sakhaee, Li,
  Basart, Li, Forsyth, and Hendrycks]{Mazeika2024HarmBench}
Mantas Mazeika, Long Phan, Xuwang Yin, Andy Zou, Zifan Wang, Norman Mu, Elham
  Sakhaee, Nathaniel Li, Steven Basart, Bo~Li, David Forsyth, and Dan
  Hendrycks.
\newblock {HarmBench}: A standardized evaluation framework for automated red
  teaming and robust refusal.
\newblock In \emph{Proceedings of the 41st International Conference on Machine
  Learning}, volume 235 of \emph{Proceedings of Machine Learning Research},
  pages 35181--35224. PMLR, 2024.
\newblock URL \url{https://proceedings.mlr.press/v235/mazeika24a.html}.

\bibitem[Parrish et~al.(2022)Parrish, Chen, Nangia, Padmakumar, Phang,
  Thompson, Htut, and Bowman]{Parrish2022BBQ}
Alicia Parrish, Angelica Chen, Nikita Nangia, Vishakh Padmakumar, Jason Phang,
  Jana Thompson, Phu~Mon Htut, and Samuel Bowman.
\newblock {BBQ}: A hand-built bias benchmark for question answering.
\newblock In \emph{Findings of the Association for Computational Linguistics:
  ACL 2022}, pages 2086--2105. Association for Computational Linguistics, 2022.
\newblock \doi{10.18653/v1/2022.findings-acl.165}.
\newblock URL \url{https://aclanthology.org/2022.findings-acl.165/}.

\bibitem[Pearl(2014)]{Pearl2014Simpson}
Judea Pearl.
\newblock Comment: Understanding {Simpson}'s paradox.
\newblock \emph{The American Statistician}, 68\penalty0 (1):\penalty0 8--13,
  2014.

\bibitem[Proskurina et~al.(2024)Proskurina, Brun, Metzler, and
  Velcin]{Proskurina2024QuantConfidence}
Irina Proskurina, Luc Brun, Guillaume Metzler, and Julien Velcin.
\newblock When quantization affects confidence of large language models?
\newblock \emph{Findings of the Association for Computational Linguistics:
  NAACL 2024}, pages 1918--1928, 2024.
\newblock \doi{10.18653/v1/2024.findings-naacl.124}.
\newblock URL \url{https://aclanthology.org/2024.findings-naacl.124/}.

\bibitem[Qi et~al.(2025)Qi, Panda, Lyu, Ma, Roy, Beirami, Mittal, and
  Henderson]{Qi2025FewTokensDeep}
Xiangyu Qi, Ashwinee Panda, Kaifeng Lyu, Xiao Ma, Subhrajit Roy, Ahmad Beirami,
  Prateek Mittal, and Peter Henderson.
\newblock Safety alignment should be made more than just a few tokens deep.
\newblock In \emph{The Thirteenth International Conference on Learning
  Representations}, 2025.
\newblock URL \url{https://openreview.net/forum?id=6Mxhg9PtDE}.

\bibitem[Ren et~al.(2024)Ren, Basart, Khoja, Pan, Gatti, Phan, Yin, Mazeika,
  Mukobi, Kim, Fitz, and Hendrycks]{Ren2024Safetywashing}
Richard Ren, Steven Basart, Adam Khoja, Alexander Pan, Alice Gatti, Long Phan,
  Xuwang Yin, Mantas Mazeika, Gabriel Mukobi, Ryan~Hwang Kim, Stephen Fitz, and
  Dan Hendrycks.
\newblock Safetywashing: Do {AI} safety benchmarks actually measure safety
  progress?
\newblock In \emph{Advances in Neural Information Processing Systems 37
  (Datasets and Benchmarks Track)}, 2024.
\newblock \doi{10.52202/079017-2190}.
\newblock URL
  \url{https://proceedings.neurips.cc/paper_files/paper/2024/hash/7ebcdd0de471c027e67a11959c666d74-Abstract-Datasets_and_Benchmarks_Track.html}.

\bibitem[Salaudeen et~al.(2025)Salaudeen, Reuel, Ahmed, Bedi, Robertson,
  Sundar, Domingue, Wang, and Koyejo]{Salaudeen2025MeasurementMeaning}
Olawale Salaudeen, Anka Reuel, Ahmed Ahmed, Suhana Bedi, Zachary Robertson,
  Sudharsan Sundar, Ben Domingue, Angelina Wang, and Sanmi Koyejo.
\newblock Measurement to meaning: A validity-centered framework for {AI}
  evaluation.
\newblock \emph{arXiv preprint arXiv:2505.10573}, 2025.
\newblock URL \url{https://arxiv.org/abs/2505.10573}.

\bibitem[Simpson(1951)]{Simpson1951}
Edward~H. Simpson.
\newblock The interpretation of interaction in contingency tables.
\newblock \emph{Journal of the Royal Statistical Society: Series B
  (Methodological)}, 13\penalty0 (2):\penalty0 238--241, 1951.

\bibitem[Tan et~al.(2026)Tan, Song, Cheng, Liu, Zhai, Hong, Wang, Xiang, and
  Yuan]{Tan2026QRealign}
Qitao Tan, Xiaoying Song, Ningxi Cheng, Ninghao Liu, Xiaoming Zhai, Lingzi
  Hong, Yanzhi Wang, Zhen Xiang, and Geng Yuan.
\newblock Q-realign: Piggybacking realignment on quantization for safe and
  efficient {LLM} deployment.
\newblock \emph{arXiv preprint arXiv:2601.08089}, 2026.
\newblock URL \url{https://arxiv.org/abs/2601.08089}.

\bibitem[Truong et~al.(2025)Truong, Tu, Liang, Li, and
  Koyejo]{Truong2025AmortizedEval}
Sang Truong, Yuheng Tu, Percy Liang, Bo~Li, and Sanmi Koyejo.
\newblock Reliable and efficient amortized model-based evaluation.
\newblock \emph{arXiv preprint arXiv:2503.13335}, 2025.
\newblock URL \url{https://arxiv.org/abs/2503.13335}.

\bibitem[Wee et~al.(2025)Wee, Kim, Kim, Hwang, and
  Kwak]{Wee2025SafetyPreservingPTQ}
Sunghyun Wee, Suyoung Kim, Hyeonjin Kim, Kyomin Hwang, and Nojun Kwak.
\newblock Safety-preserving {PTQ} via contrastive alignment loss.
\newblock \emph{arXiv preprint arXiv:2511.07842}, 2025.
\newblock URL \url{https://arxiv.org/abs/2511.07842}.

\bibitem[Wei et~al.(2024)Wei, Huang, Huang, Xie, Qi, Xia, Mittal, Wang, and
  Henderson]{Wei2024BrittlenessAlignment}
Boyi Wei, Kaixuan Huang, Yangsibo Huang, Tinghao Xie, Xiangyu Qi, Mengzhou Xia,
  Prateek Mittal, Mengdi Wang, and Peter Henderson.
\newblock Assessing the brittleness of safety alignment via pruning and
  low-rank modifications.
\newblock In \emph{Proceedings of the 41st International Conference on Machine
  Learning}, volume 235 of \emph{Proceedings of Machine Learning Research},
  pages 52588--52610. PMLR, 2024.
\newblock URL \url{https://proceedings.mlr.press/v235/wei24f.html}.

\bibitem[Weidinger et~al.(2024)Weidinger, Barnhart, Brennan, Butterfield,
  Young, Hawkins, Hendricks, Comanescu, Chang, Rodriguez, Beroshi, Bloxwich,
  Proleev, Chen, Farquhar, Ho, Gabriel, Dafoe, and
  Isaac]{Weidinger2024HolisticSafety}
Laura Weidinger, Joslyn Barnhart, Jenny Brennan, Christina Butterfield, Susie
  Young, Will Hawkins, Lisa~Anne Hendricks, Ramona Comanescu, Oscar Chang,
  Mikel Rodriguez, Jennifer Beroshi, Dawn Bloxwich, Lev Proleev, Jilin Chen,
  Sebastian Farquhar, Lewis Ho, Iason Gabriel, Allan Dafoe, and William Isaac.
\newblock Holistic safety and responsibility evaluations of advanced {AI}
  models.
\newblock \emph{arXiv preprint arXiv:2404.14068}, 2024.
\newblock URL \url{https://arxiv.org/abs/2404.14068}.

\bibitem[Xiao et~al.(2023)Xiao, Lin, Seznec, Wu, Demouth, and
  Han]{Xiao2023SmoothQuant}
Guangxuan Xiao, Ji~Lin, Mickael Seznec, Hao Wu, Julien Demouth, and Song Han.
\newblock {SmoothQuant}: Accurate and efficient post-training quantization for
  large language models.
\newblock In \emph{Proceedings of the 40th International Conference on Machine
  Learning}, volume 202 of \emph{Proceedings of Machine Learning Research},
  pages 38087--38099. PMLR, 2023.
\newblock URL \url{https://proceedings.mlr.press/v202/xiao23c.html}.

\bibitem[Xu et~al.(2024)Xu, Gupta, Li, Bentham, and
  Srikumar]{Xu2024BeyondPerplexity}
Zhichao Xu, Ashim Gupta, Tao Li, Oliver Bentham, and Vivek Srikumar.
\newblock Beyond perplexity: Multi-dimensional safety evaluation of {LLM}
  compression.
\newblock In \emph{Findings of the Association for Computational Linguistics:
  EMNLP 2024}, pages 15359--15396. Association for Computational Linguistics,
  2024.
\newblock \doi{10.18653/v1/2024.findings-emnlp.901}.
\newblock URL \url{https://aclanthology.org/2024.findings-emnlp.901/}.

\bibitem[Zhang et~al.(2020)Zhang, Kishore, Wu, Weinberger, and
  Artzi]{Zhang2020BERTScore}
Tianyi Zhang, Varsha Kishore, Felix Wu, Kilian~Q. Weinberger, and Yoav Artzi.
\newblock {BERTScore}: Evaluating text generation with {BERT}.
\newblock In \emph{International Conference on Learning Representations}, 2020.
\newblock URL \url{https://openreview.net/forum?id=SkeHuCVFDr}.

\bibitem[Zheng et~al.(2023)Zheng, Chiang, Sheng, Zhuang, Wu, Zhuang, Lin, Li,
  Li, Xing, Zhang, Gonzalez, and Stoica]{Zheng2023MTBench}
Lianmin Zheng, Wei-Lin Chiang, Ying Sheng, Siyuan Zhuang, Zhanghao Wu, Yonghao
  Zhuang, Zi~Lin, Zhuohan Li, Dacheng Li, Eric Xing, Hao Zhang, Joseph~E.
  Gonzalez, and Ion Stoica.
\newblock Judging {LLM}-as-a-judge with {MT}-bench and chatbot arena.
\newblock In \emph{Advances in Neural Information Processing Systems 36
  (Datasets and Benchmarks Track)}, 2023.
\newblock URL
  \url{https://proceedings.neurips.cc/paper_files/paper/2023/hash/91f18a1287b398d378ef22505bf41832-Abstract-Datasets_and_Benchmarks.html}.

\bibitem[Zhu et~al.(2024)Zhu, Li, Liu, Ma, and Wang]{Zhu2024SurveyCompression}
Xunyu Zhu, Jian Li, Yong Liu, Can Ma, and Weiping Wang.
\newblock A survey on model compression for large language models.
\newblock \emph{Transactions of the Association for Computational Linguistics},
  12:\penalty0 1556--1577, 2024.
\newblock \doi{10.1162/tacl_a_00704}.
\newblock URL \url{https://aclanthology.org/2024.tacl-1.85/}.

\bibitem[Zou et~al.(2023)Zou, Wang, Carlini, Nasr, Kolter, and
  Fredrikson]{Zou2023AdvBench}
Andy Zou, Zifan Wang, Nicholas Carlini, Milad Nasr, J.~Zico Kolter, and Matt
  Fredrikson.
\newblock Universal and transferable adversarial attacks on aligned language
  models.
\newblock \emph{arXiv preprint arXiv:2307.15043}, 2023.
\newblock URL \url{https://arxiv.org/abs/2307.15043}.

\end{thebibliography}

\appendix
\section{Ethical Considerations, Data Availability, and Disclosures}

\subsection{Ethical Considerations}

This paper analyzes quality and safety behavior under quantization using stored evaluation outputs, AWQ/GPTQ checkpoints, and judge annotations. It contains no human-subjects component and no private data. Because the safety side inherits harmful-prompt evaluation from the benchmark tasks, the public release prioritizes scored outputs, matrices, and build scripts over redistribution of packaged harmful-request batteries.

\subsection{Data Availability}

The reproduction package for this paper includes:
\begin{itemize}[leftmargin=*]
  \item the 51-row study matrix and the file-level lineage needed to rebuild it;
  \item the quality, direct-safety, and primary-judge outputs required to reproduce the paper tables;
  \item the AWQ/GPTQ expansion outputs used in the cross-method comparison;
  \item the second-judge robustness outputs, including the stratified sample, combined Claude outputs, and agreement summary; and
  \item the four-phase mechanistic follow-up outputs for the 17-cell HF-backed subset, including the completed phase-4 7B rerun results; and
  \item build scripts and audit notes sufficient to regenerate the manuscript assets.
\end{itemize}

\noindent File lineage is preserved in the supplementary materials together with the overlap tables and build scripts required to reproduce the reported summary statistics.

\noindent This is a reproduction package for the scoped quality--safety matrix, not a general-purpose benchmark dataset. It does not include closed-source baselines, a public leaderboard workload, or the breadth needed to rank models outside the quantization cells studied in the paper.

\subsection{Funding}

No external funding was received for this work.

\subsection{Competing Interests}

The authors declare no competing interests.

\subsection{Disclosure}

This manuscript combines prior GGUF evaluations with newly completed quantization, evaluation, and second-judge robustness runs. No claims in the manuscript rely on private evaluations or hidden data sources outside the documented file lineage.

\subsection{Compute Resources}
\label{subsec:compute_resources}

Small-model inference (llama3.2-1b, llama3.2-3b, qwen2.5-1.5b, phi-2) was run on a consumer laptop with an NVIDIA GeForce RTX 4080 Laptop GPU (12\,GB VRAM). The 7B models (qwen2.5-7b, mistral-7b) used a RunPod instance with an NVIDIA RTX 6000 Ada (48\,GB VRAM) at \$0.77/hr for quantization and evaluation. GGUF inference used llama.cpp via Ollama; AWQ checkpoints used AutoAWQ; GPTQ checkpoints used AutoGPTQ. The primary judge (gemma3:12b) ran locally via Ollama on the RTX 4080 Laptop GPU.

The mechanistic follow-up used the same hardware split. Phases 1--3 and the small-model phase-4 cells ran locally as forward-pass-only probes on the RTX 4080 Laptop GPU. The four 7B phase-4 cells (qwen2.5-7b AWQ/GPTQ and mistral-7b AWQ/GPTQ) were completed on the RTX 6000 Ada instance because simultaneous FP16-anchor and quantized hidden-state extraction exceeded the local VRAM budget.

Approximate compute breakdown:
\begin{itemize}[leftmargin=*,itemsep=2pt]
  \item Quality evaluation (41,895 generations across 51 cells): ${\sim}$40 GPU-hours
  \item Safety evaluation (48,603 generations): ${\sim}$35 GPU-hours
  \item Primary judge (21,096 labels): ${\sim}$15 GPU-hours
  \item AWQ/GPTQ quantization (11 checkpoints): ${\sim}$12 GPU-hours (RunPod)
  \item Second-judge pass (11,470 rows via Claude Sonnet 4 API): ${\sim}$\$25 API cost
\end{itemize}

\noindent Total project compute, including preliminary experiments not reported in the paper, was approximately 150 GPU-hours and \$25 in API costs.

\subsection{Pipeline Overview}

\begin{figure}[ht]
  \centering
  \includegraphics[width=\textwidth]{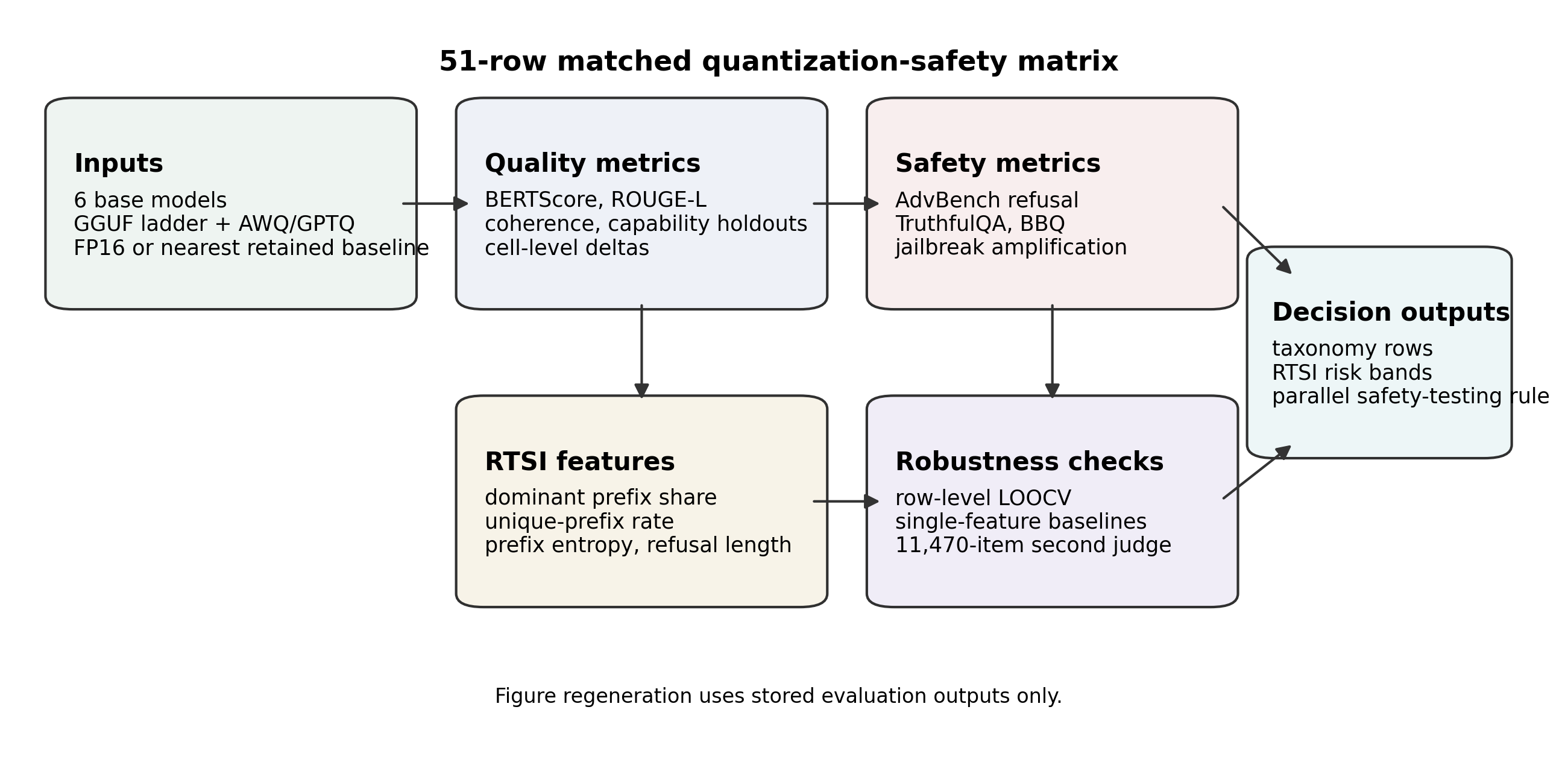}
  \caption{Paper pipeline for the completed study dataset. GGUF ladder rows and new AWQ/GPTQ runs are normalized into a 51-row quality--safety matrix. Direct safety metrics, gemma3:12b judge summaries, deployment taxonomy, and the Claude Sonnet 4 robustness pass are all derived from the same row set.}
  \label{fig:design_appendix}
\end{figure}

\clearpage
\subsection{Supplementary Result Tables and Figures}

\begin{table}[ht]
\centering
\caption{Evaluation slices used in the matched matrix. Prompt budgets are per model--quantization cell. The direct-safety aggregates use only the four safety-specific tasks; retained MMLU and ARC slices are carried through the raw safety runs as capability holdouts and are not folded into refusal, truthfulness, or bias scores.}
\label{tab:task_slices_appendix}
\small
\begin{tabularx}{\textwidth}{>{\raggedright\arraybackslash}p{0.18\textwidth} >{\raggedright\arraybackslash}X >{\raggedleft\arraybackslash}p{0.12\textwidth} >{\raggedright\arraybackslash}p{0.24\textwidth}}
\toprule
\textbf{Slice} & \textbf{Tasks} & \textbf{Nominal prompts / cell} & \textbf{Used for} \\
\midrule
Quality & summarization, QA, code generation, creative writing, classification, MMLU, ARC & 735 & BERTScore, ROUGE-L, coherence, accuracy, length, repetition \\
Direct safety & AdvBench refusal, jailbreak amplification, TruthfulQA, BBQ & 468 & refusal, truthfulness, bias resistance \\
Retained holdouts & MMLU, ARC within the raw safety runs & 485 & capability holdouts only; excluded from direct-safety aggregates \\
Primary judge & same 4 safety tasks as direct safety & 468 & judge refusal, judge truthfulness, judge bias resistance; observed judged rows vary after filtering \\
\bottomrule
\end{tabularx}
\end{table}

\begin{table}[ht]
\centering
\caption{Four-probe mechanistic follow-up on the 17-cell Hugging Face-backed FP16/AWQ/GPTQ subset. The first three probes are weak or null predictors of dangerous rows; the safety-associated-neuron probe shows a real overall effect without becoming regime-specific.}
\label{tab:mechanistic_followup_appendix}
\small
\begin{tabularx}{\textwidth}{>{\raggedright\arraybackslash}p{0.23\textwidth} >{\raggedright\arraybackslash}X >{\raggedright\arraybackslash}p{0.24\textwidth}}
\toprule
\textbf{Probe} & \textbf{Main result} & \textbf{Deployment value} \\
\midrule
Entropy shift & RTSI correlation \(r = 0.083\), regime \(p = 0.606\) & no usable screen \\
Refusal direction geometry & cosine \(> 0.97\) in every cell; RTSI \(r = -0.144\), regime \(p = 0.606\) & no usable screen \\
Calibration drift & all \(|r| < 0.09\), all regime \(p \ge 0.788\) & no usable screen \\
Safety-associated-neuron quantization error & mean safety/non-safety ratio \(= 1.395\times\), \(p = 4.89 \times 10^{-7}\); regime \(p = 0.979\) & explanatory only \\
RTSI & low-risk bucket \(= 23/45\); danger recall \(= 1.0\) in-sample and under LOOCV & conservative study-internal screen \\
\bottomrule
\end{tabularx}
\end{table}

\begin{figure}[ht]
  \centering
  \includegraphics[width=0.78\textwidth]{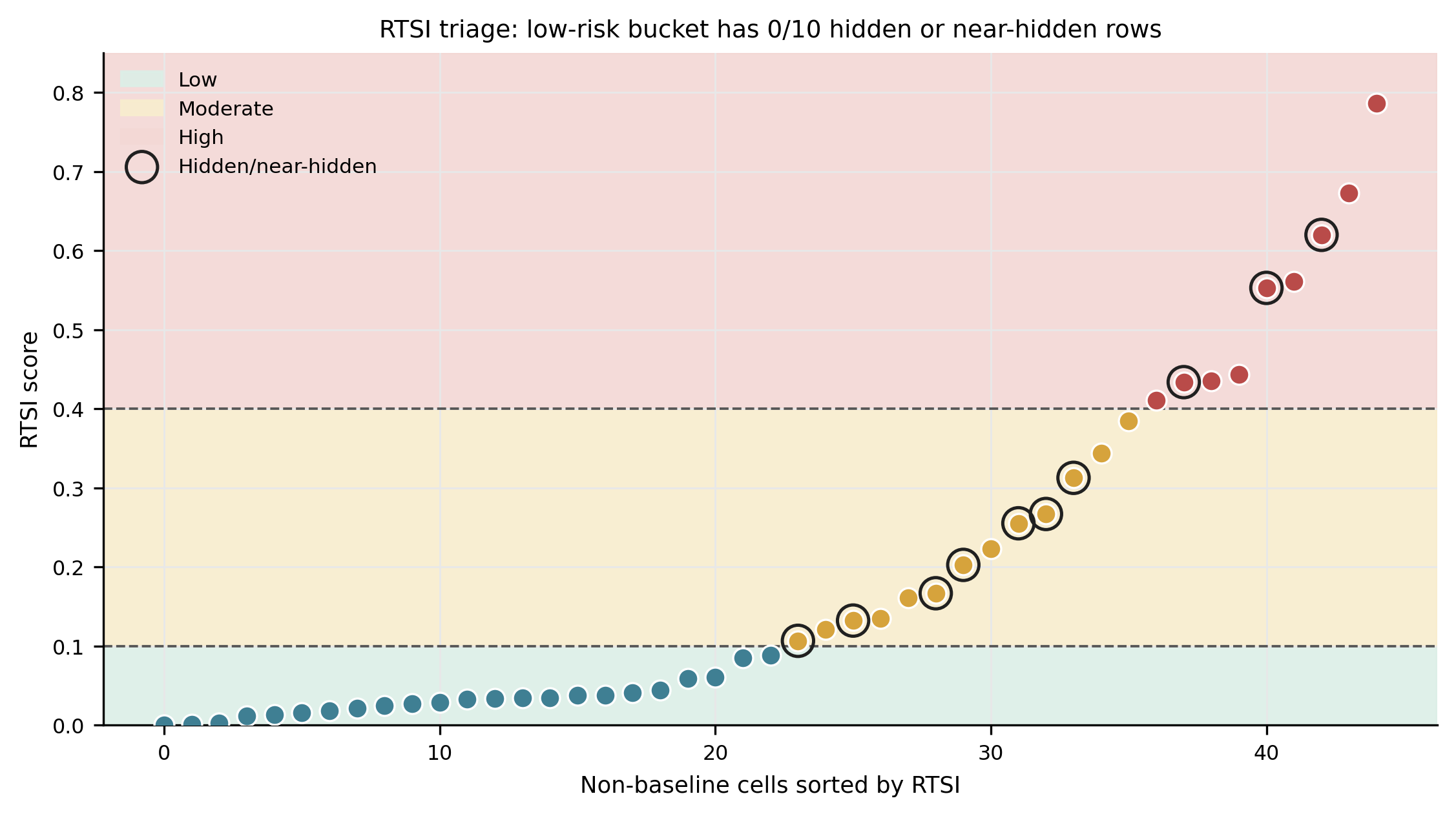}
  \caption{RTSI as a conservative study-internal deployment screen. Scores below 0.10 form the low-risk bucket; scores from 0.10 to 0.40 are moderate risk; scores at or above 0.40 are high risk. On the final matrix, the low-risk bucket contains 23/45 non-baseline rows and 0/10 hidden- or near-hidden-danger rows, while every AWQ/GPTQ row lands outside the low-risk tier. The same low-risk count and zero-false-negative property hold under row-level leave-one-out validation.}
  \label{fig:mitigator_appendix}
\end{figure}

\begin{figure}[ht]
  \centering
  \includegraphics[width=0.78\textwidth]{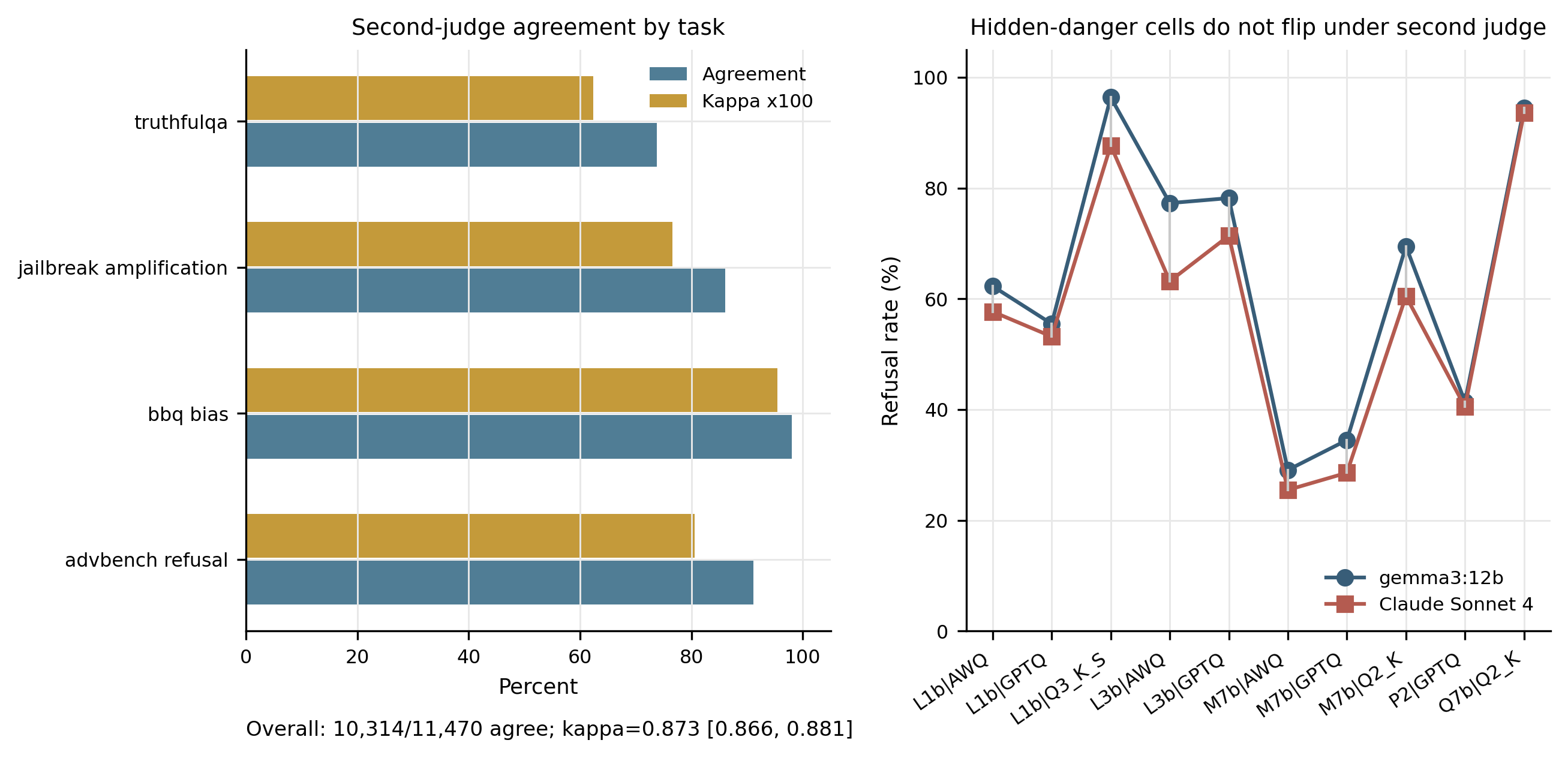}
\caption{Second-judge robustness on the predefined 11,470-row stratified second-judge set. Claude Sonnet 4 agrees strongly with the primary gemma3:12b judge, and no hidden-danger or near-hidden cell flips regime.}
  \label{fig:judge_robustness_appendix}
\end{figure}

\clearpage
\subsection{RTSI Reporting Card}
\label{app:rtsi-reporting-card}

The reporting card below is the canonical schema for citing RTSI in a quant-safety claim. Example values illustrate a hidden-danger row from the matched 51-cell matrix.

\begin{verbatim}
# RTSI reporting card
# calibration: 51-cell matrix; 45 non-baseline rows
cell_identity:
  model_family:           llama-3.2
  base_model:             llama3.2-1b
  quantization_format:    GPTQ-INT4
  judge_model:            gemma3:12b           # primary
  prompt_battery:         [AdvBench, TruthfulQA, BBQ, jailbreak-amp]
  harm_categories:        [harmful-request, jailbreak, truthfulness, bias]

calibration_provenance:
  n_baseline_rows:        45                   # non-baseline rows
  n_total_cells:          51

feature_deltas:
  # raw deltas vs FP16; normalized internally
  delta_dominant_prefix_share:    -0.335
  delta_unique_prefix_rate:       +0.405
  delta_prefix_entropy_norm:      +0.319
  delta_mean_refusal_token_length:+67.89

score:
  rtsi:                   0.43
  classification:         HIGH
  bands:
    LOW < 0.10
    MODERATE 0.10-0.40
    HIGH >= 0.40

loocv_check:
  loocv_recall_on_calibration: 10/10
  is_recalibrated:        false
\end{verbatim}

Compute RTSI whenever retained quality (BERTScore, ROUGE-L, accuracy, judged coherence) is being used as a deployment proxy without a matched direct-safety battery on the same cells. The default cutoffs (\textsc{Low} $<0.10$, \textsc{Moderate} $0.10$--$0.40$, \textsc{High} $\geq 0.40$) are calibrated on quantization perturbations (GGUF k-quants, AWQ, GPTQ) of small-to-mid open-weights models (1B--7B) under a single LLM judge or a triangulated pair with $\kappa \geq 0.7$. Recalibrate the four feature weights and min--max ranges before citing RTSI when the judge family changes, the model scale crosses 13B+, or the prompt battery is multi-turn or non-English; report \texttt{is\_recalibrated: true} and the new \texttt{n\_baseline\_rows} in that case.

\begin{quote}
\textbf{When citing RTSI}, report all six identity fields plus the four feature deltas, the RTSI scalar, the classification band (\textsc{Low}/\textsc{Moderate}/\textsc{High}), and the LOOCV-recall on the original calibration corpus (e.g., \texttt{RTSI=0.43 / HIGH / 10/10 LOOCV} on the cell tuple \texttt{(llama3.2-1b, GPTQ-INT4, gemma3:12b, AdvBench+TruthfulQA+BBQ+jailbreak-amp)}).
\end{quote}

\subsection{Computing RTSI on Arbitrary Quantization-Safety Data}
\label{app:rtsi-csv}

The reference implementation ships a generic CSV ingest mode for retrofitting RTSI to any quantization-safety study with refusal-template features per cell. The CSV must contain \texttt{cell\_id}, \texttt{is\_baseline}, \texttt{dominant\_prefix\_share}, \texttt{unique\_prefix\_rate}, \texttt{prefix\_entropy\_norm}, and \texttt{mean\_refusal\_token\_length}. The \texttt{cell\_id} may be formatted as \texttt{model\_id\_\_variant} or paired with an explicit \texttt{model\_id} column. Each model must have exactly one baseline row; non-baseline rows are scored against the matched baseline.

\begin{verbatim}
python compute_rtsi_csv.py \
    --input my_refusal_features.csv \
    --output rtsi_results.json
# CSV columns:
# cell_id, is_baseline, dominant_prefix_share,
# unique_prefix_rate, prefix_entropy_norm,
# mean_refusal_token_length
\end{verbatim}

The script groups by model, computes per-feature deltas vs the matched baseline, applies the matrix-calibrated \texttt{RTSI\_WEIGHTS}, and emits per-cell JSON with \texttt{rtsi}, \texttt{classification}, \texttt{feature\_deltas}, and any \texttt{implausible\_flags} (e.g., \texttt{unique\_prefix\_rate}~$>1$, negative entropy). Cells with no baseline or multiple baselines are reported as \texttt{insufficient\_data}. The same code path produces every RTSI number in this paper. A self-test (\texttt{python compute\_rtsi\_csv.py --self-test}) covers \textsc{Low}, \textsc{Moderate}, \textsc{High}, missing-baseline, and implausible-feature cases.

\subsection{External Comparative Anchor: Public Refusal-Completion Releases}
\label{app:rtsi-external-anchor}

To position the three-band classification on the scale of public quant-safety releases, we attempted to retroactively compute RTSI on externally-published refusal data. The required data shape is raw refusal completions across compressed variants of a fixed base model on a shared prompt set, so the four refusal-template features can be extracted per (model, variant) cell. No current public release provides exactly this combination.

\begin{table}[h]
  \centering
  \footnotesize
  \setlength{\tabcolsep}{4pt}
  \caption{External comparative anchor for RTSI. JailbreakBench JBC artifacts release raw refusal completions across model pairs but use a cross-base-model design (Llama-2-7B vs Vicuna-13B) rather than within-base compressed variants, which is a category mismatch for RTSI as defined. Other public quant-safety releases either publish summary scores only or have not yet released raw completions.}
  \label{tab:rtsi-external-anchor}
  \begin{tabular}{>{\raggedright\arraybackslash}p{0.32\linewidth}>{\raggedright\arraybackslash}p{0.34\linewidth}rl}
    \toprule
    Source & Model pair / availability & RTSI & Classification \\
    \midrule
    This paper (llama3.2-1b GPTQ vs FP16) & within-base quant variant pair, $n=51$ cells, 100 prompts/cell & 0.43 & HIGH \\
    JailbreakBench JBC \cite{Chao2024JailbreakBench} & cross-base-model: Llama-2-7B vs Vicuna-13B (RTSI-inapplicable) & --- & --- \\
    Compressed Trust \cite{Hong2024CompressedTrust} & raw refusal completions not yet public on the dataset hub & --- & --- \\
    HarmBench \cite{Mazeika2024HarmBench} / BeaverTails \cite{Ji2023BeaverTails} & summary scores released, raw completions not in the public artifact & --- & --- \\
    \bottomrule
  \end{tabular}
\end{table}

\paragraph{What the JBB-JBC degeneracy shows.} Running the four-feature extractor over the JBC artifacts at \texttt{JailbreakBench/artifacts/main/attack-artifacts/JBC/manual/} produces a degenerate result---Llama-2-7B-chat returns 79/100 refusals all opening with the single prefix ``i cannot fulfill your request'' (\texttt{dominant\_prefix\_share}~$=1.0$, \texttt{unique\_prefix\_rate}~$=1/79$), and Vicuna-13B-v1.5 returns 0/100 refusals because every response opens with \texttt{Niccolo:} under the AIM role-play template. RTSI computes as \textsc{Low} ($\textit{rtsi}=0.0$) with all four feature deltas flagged as missing/degenerate (the per-feature inputs are undefined since Vicuna emits 0/100 refusals); we therefore report the cell as effectively \texttt{insufficient\_data} for comparative-anchor purposes. The surface artifact is the AIM template; the underlying gap is structural: JBB pairs models with different post-training pipelines (RLHF-aligned Llama-2 vs ShareGPT-SFT Vicuna), which is a category mismatch for an index defined over compressed variants of a fixed base. The honest comparative gap RTSI fills is therefore: \emph{no current public release exposes raw refusal completions across compressed variants of a fixed base model on a shared prompt set}; this paper's matrix is the niche such a release would occupy.

\section*{Broader Impact, Ethics, and Data / Code Availability}
\label{sec:broader_impact}

\subsection*{Broader Impact}

The positive externality of this work is a scoped proxy-validity audit that
tells deployment teams not to substitute retained quality metrics for direct
safety evaluation when shipping GGUF, AWQ, or GPTQ checkpoints on the model
families studied here (\S\ref{sec:danger}, \S\ref{sec:results}). The negative
externality we flag is the risk of the Refusal Template Stability Index (RTSI)
being repurposed as a general ``safety-passed'' rubber stamp outside the
51-row matrix on which it was calibrated; we therefore present RTSI as a
calibrated behavioral screen on this matrix---validated by row-level leave-one-out
but not against held-out feature sets or thresholds---and explicitly route
all hidden- and near-hidden-danger rows to direct safety evaluation
(\S\ref{sec:mitigator}, \S\ref{sec:gating}). The findings are a matched
proxy-validity audit over the measured cells (\S\ref{sec:discussion}).

\subsection*{Ethics}

This paper evaluates publicly released harmful-prompt suites (AdvBench,
TruthfulQA, BBQ, and the jailbreak-amplification workload described in
\S\ref{sec:design}) against open-weight models and does not introduce any
novel harmful content. API-side evaluation of the Claude Sonnet 4 second
judge was performed through a researcher-credit account with pre-approved
safety-evaluation traffic, and only aggregated refusal, truthfulness, and
bias-resistance statistics are released; no verbatim harmful completions
leave the local evaluation environment. The work involves no human subjects,
no crowdsourcing, and no personal data, so no IRB review was required. The
LLM-judge pipeline is methodological infrastructure, not a human-subjects
instrument. Responsible-use expectations for the reproduction package are
stated in \S\ref{sec:broader_impact} (this appendix) and in the data-release
clause of the submission materials appendix.

\subsection*{Data and Code Availability}

The reproduction package for this paper includes the 51-row
study matrix, the quality / direct-safety / primary-judge outputs, the
AWQ/GPTQ expansion outputs, the Claude Sonnet 4 second-judge outputs
over the predefined 11{,}470-row stratified sample, the four-phase
mechanistic follow-up outputs on the 17-cell HF-backed subset, and the
build scripts needed to regenerate the manuscript assets. This is a
scoped proxy-validity audit artifact, not a community benchmark dataset with
closed-source baselines or a public leaderboard. Public dataset and
code-repository URLs should be added when the release archive is posted;
review artifacts are supplied through the conference
supplementary-material channel, as noted in the NeurIPS 2026 checklist
appendix (\S\ref{sec:neurips_checklist}).

\section*{NeurIPS 2026 Paper Checklist}
\label{sec:neurips_checklist}

\begin{enumerate}
  \item \textbf{Claims.} Do the main claims made in the abstract and
  introduction accurately reflect the paper's contributions and scope?
  \textbf{Answer:} Yes. \emph{Justification:} The abstract and
  \S\ref{sec:intro} describe the study as a scoped proxy-validity audit over a
  matched 51-row matrix, and every headline number (36/36 sign splits, 9
  hidden-danger + 1 near-hidden row, 7/11 AWQ/GPTQ rows, RTSI low-risk
  bucket 23/45, second-judge $\kappa = 0.873$) is anchored in
  \S\ref{sec:results}.

  \item \textbf{Limitations.} Does the paper discuss the limitations of
  the work performed by the authors?
  \textbf{Answer:} Yes. \emph{Justification:} \S\ref{sec:discussion}
  enumerates the 6-model / 4-family / $\leq$7B coverage limitation, the
  within-stack conclusion, the judge-family overlap, the RTSI
  row-level-LOOCV-only calibration, and the benchmark-breadth future work.

  \item \textbf{Theory assumptions and proofs.} For each theoretical
  result, does the paper provide the full set of assumptions and a
  complete (and correct) proof?
  \textbf{Answer:} NA. \emph{Justification:} The paper contains no
  theoretical theorems; all claims are empirical over the 51-row matrix.

  \item \textbf{Experimental result reproducibility.} Does the paper
  fully disclose all the information needed to reproduce the main
  experimental results to the extent that it affects the main claims?
  \textbf{Answer:} Partial. \emph{Justification:} \S\ref{sec:design}
  documents the matched-matrix construction, task slices, quantization
  pipeline, and judge configuration, and the reproduction package (via
  \texttt{scripts/build\_reproducibility\_bundle.py}) ships aggregated
  labels, second-judge outputs, and analysis scripts sufficient to
  regenerate every headline number; raw harmful-prompt batteries and
  public dataset/code URLs are withheld during review and are
  available via the conference supplementary-material channel.

  \item \textbf{Open access to data and code.} Does the paper provide
  open access to data and code, with sufficient instructions to
  faithfully reproduce the main experimental results?
  \textbf{Answer:} Partial. \emph{Justification:} Aggregated evaluation
  outputs, second-judge relabellings, mechanistic-probe statistics, and
  the analysis harness are released in the reproducibility bundle
  described in \S\ref{sec:broader_impact}; verbatim harmful completions
  and the fully packaged attack batteries are withheld under the
  responsible-use posture, and public dataset/code URLs are omitted
  during review (review artifacts are supplied through the
  conference supplementary-material channel).

  \item \textbf{Experimental setting/details.} Does the paper specify
  all training and test details (splits, hyperparameters, optimizer,
  etc.) necessary to understand the results?
  \textbf{Answer:} Yes. \emph{Justification:} \S\ref{sec:design} and
  \S\ref{sec:metrics} specify task prompts per cell, the GGUF ladder, AWQ
  and GPTQ INT4 configurations, decoding settings, and judge prompts; the
  submission-materials appendix lists file lineage.

  \item \textbf{Experiment statistical significance.} Does the paper
  report error bars or confidence intervals or statistical significance
  tests?
  \textbf{Answer:} Yes. \emph{Justification:} \S\ref{sec:results} reports
  second-judge Cohen's $\kappa$ with a 95\% bootstrap CI, the safety-neuron
  quantization-error effect with $p = 4.89 \times 10^{-7}$, and
  sign-heterogeneity across all 36 quality--safety pairings; RTSI
  screening is validated under row-level LOOCV.

  \item \textbf{Experiments compute resources.} Does the paper provide
  sufficient compute detail to reproduce the experiments?
  \textbf{Answer:} Yes. \emph{Justification:} \S\ref{subsec:compute_resources}
  in the submission-materials appendix lists the RTX 4080 Laptop + RTX 6000
  Ada split, the per-phase GPU-hours (40 + 35 + 15 + 12 GPU-hours), and the
  approximate \$25 Claude Sonnet 4 API cost.

  \item \textbf{Code of ethics.} Does the research conform with the
  NeurIPS Code of Ethics in every respect?
  \textbf{Answer:} Yes. \emph{Justification:} The Ethics subsection in
  \S\ref{sec:broader_impact} documents the use of publicly released
  harmful-prompt suites without introducing new harmful content,
  researcher-credit-account API usage, and the absence of human subjects.

  \item \textbf{Broader impacts.} Does the paper discuss both potential
  positive and negative societal impacts?
  \textbf{Answer:} Yes. \emph{Justification:} The Broader Impact subsection
  in \S\ref{sec:broader_impact} names the positive outcome (deployment
  triage against quality-as-safety-proxy substitution) and the negative
  risk (RTSI misuse as a general ``safety-passed'' stamp outside the
  studied matrix).

  \item \textbf{Safeguards.} Does the paper describe safeguards for
  responsible release of high-misuse-risk data or models?
  \textbf{Answer:} Yes. \emph{Justification:} \S\ref{sec:broader_impact}
  and the submission-materials appendix state that the reproduction
  package releases aggregated refusal, truthfulness, and bias-resistance
  statistics only; no verbatim harmful completions are redistributed, and
  harmful-prompt evaluation is inherited from public benchmarks rather
  than repackaged.

  \item \textbf{Licenses for existing assets.} Are creators/original
  owners of used assets properly credited with license and terms of use?
  \textbf{Answer:} Yes. \emph{Justification:} \texttt{refs.bib} cites the
  AdvBench, TruthfulQA, BBQ, BERTScore, ROUGE, AutoAWQ, AutoGPTQ, and
  \texttt{llama.cpp}/Ollama sources used in the pipeline described in
  \S\ref{sec:design}; model-card license terms are inherited from the
  upstream Hugging Face repositories.

  \item \textbf{New assets.} Are new assets introduced in the paper
  well documented?
  \textbf{Answer:} Yes. \emph{Justification:} The new assets are the
  51-row quality--safety matrix, the RTSI screening heuristic, and the four-probe
  mechanistic follow-up outputs; all three are described in
  \S\ref{sec:design}, \S\ref{sec:mitigator}, and the submission-materials
  appendix, and are packaged by the reproducibility bundle builder cited
  in \S\ref{sec:broader_impact}.

  \item \textbf{Crowdsourcing and human subjects.} For crowdsourcing
  and research with human subjects, does the paper include instructions,
  screenshots, and compensation details?
  \textbf{Answer:} NA. \emph{Justification:} No crowdsourcing or
  human-subjects data collection was performed; LLM judges
  (\texttt{gemma3:12b} primary, Claude Sonnet 4 secondary) are
  methodological infrastructure, not human raters.

  \item \textbf{IRB approvals.} Does the paper describe potential
  participant risks, disclosure, and IRB (or equivalent) approvals?
  \textbf{Answer:} NA. \emph{Justification:} No human subjects were
  involved, so no IRB review was required, as stated in the Ethics
  subsection of \S\ref{sec:broader_impact}.

  \item \textbf{LLM usage.} Does the paper declare LLM usage if it is
  an important, original, or non-standard component of the core methods?
  \textbf{Answer:} Yes. \emph{Justification:} The primary judge
  \texttt{gemma3:12b} and the Claude Sonnet 4 second judge are core
  methodological components declared in \S\ref{sec:design} and
  \S\ref{sec:results}; agreement, $\kappa$, and cell-level regime
  stability are reported explicitly.
\end{enumerate}

\end{document}